\documentclass{ieeetj}
\usepackage{cite}
\usepackage{amsmath,amssymb,amsfonts}
\usepackage{algorithmic}
\usepackage{graphicx,color}
\usepackage{textcomp}
\usepackage{xcolor}
\usepackage{hyperref}
\hypersetup{hidelinks=true}
\usepackage{algorithm,algorithmic}
\def\BibTeX{{\rm B\kern-.05em{\sc i\kern-.025em b}\kern-.08em
    T\kern-.1667em\lower.7ex\hbox{E}\kern-.125emX}}
\AtBeginDocument{\definecolor{tmlcncolor}{cmyk}{0.93,0.59,0.15,0.02}\definecolor{NavyBlue}{RGB}{0,86,125}}

\def\OJlogo{\vspace{-4pt}$<$Society logo(s) and publication title will appear here.$>$}
\def\seclogo{\vspace{10pt}$<$Society logo(s) and publication title will appear here.$>$}

\def\authorrefmark#1{\ensuremath{^{\textbf{#1}}}}

\begin{document}
\receiveddate{XX Month, XXXX}
\reviseddate{XX Month, XXXX}
\accepteddate{XX Month, XXXX}
\publisheddate{XX Month, XXXX}
\currentdate{XX Month, XXXX}
\doiinfo{XXXX.2022.1234567}

\markboth{}{Author {et al.}}

\title{An Open Toolkit for Underwater Field Robotics}

\author{Giacomo Picardi\authorrefmark{1}, Saverio Iacoponi\authorrefmark{2}, Matias Carandell\authorrefmark{3}, Jorge Aguirregomezcorta\authorrefmark{1,2}, Mrudul Chellpurath\authorrefmark{4}, Joaquin del Rio\authorrefmark{3}, Senior Member, IEEE, Marcello Calisti\authorrefmark{5} Member, IEEE, and Jacopo Aguzzi\authorrefmark{1}}
\affil{Instituto de Ciencias del Mar ICM-CSIC, Barcelona, Spain}
\affil{Department of Mechanical and Nuclear Engineering, Khalifa University, Abu Dhabi, UAE}
\affil{SARTI-MAR Research group, Universitat Politècnica de Catalunya (UPC), Barcelona Tech.; Barcelona, 08034, Spain}
\affil{KTH Royal Institute of Technology, Stockholm, Sweden}
\affil{The BioRobotics Institute, Scuola Superiore Sant'Anna, Pisa, Italy}
\corresp{Corresponding author: Giacomo Picardi (email: gpicardi@icm.csic.es).}

\begin{abstract}
Underwater robotics is becoming increasingly important for marine science, environmental monitoring, and subsea industrial operations, yet the development of underwater manipulation and actuation systems remains restricted by high costs, proprietary designs, and limited access to modular, research-oriented hardware. While open-source initiatives have democratized vehicle construction and control software, a substantial gap persists for joint-actuated systems—particularly those requiring waterproof, feedback-enabled actuation suitable for manipulators, grippers, and bioinspired devices. As a result, many research groups face lengthy development cycles, limited reproducibility, and difficulty transitioning laboratory prototypes to field-ready platforms.

To address this gap, we introduce an open, cost-effective hardware and software toolkit for underwater manipulation research. The toolkit includes a depth-rated Underwater Robotic Joint (URJ) with early leakage detection, compact control and power management electronics, and a ROS\,2-based software stack for sensing and multi-mode actuation. All CAD models, fabrication files, PCB sources, firmware, and ROS\,2 packages are openly released, enabling local manufacturing, modification, and community-driven improvement.

The toolkit has undergone extensive laboratory testing and multiple field deployments, demonstrating reliable operation up to 40 m depth across diverse applications, including a 3-DoF underwater manipulator, a tendon-driven soft gripper, and an underactuated sediment sampler. These results validate the robustness, versatility, and reusability of the toolkit for real marine environments.

By providing a fully open, field-tested platform, this work aims to lower the barrier to entry for underwater manipulation research, improve reproducibility, and accelerate innovation in underwater field robotics.
\end{abstract}



\maketitle

\section{INTRODUCTION}
Underwater operations are becoming essential in scientific and commercial sectors to support exploration, sustainable resource management, restoration and monitoring, and infrastructure maintenance in marine environments. Several global policy initiatives, including the UN Decade of Ocean Science for Sustainable Development \cite{OceanDecade2021} and the EU Marine Strategy Framework Directive \cite{MSFD2008}, emphasize the need for improved monitoring and sustainable management of marine ecosystems \cite{aguzzi2024new}. At the same time, the Blue Economy sector demands for efficient solutions to robustly and persistently operate in the proximity of submerged structures and important economical assets. As human access to underwater regions remains limited, robotics technologies play a crucial role in enabling safe and efficient operations \cite{aguzzi2022developing}. However, the development of robotic systems for underwater operations presents significant challenges due to hydrostatic pressure, electrical insulation, water ingress risks, and control complexities \cite{sivvcev2018underwater}. While propeller-driven robots—based on brushless motors with continuous rotation—benefit from a mature ecosystem of commercially available thrusters, ESC controllers, navigation modules, and open-source communities, a substantial gap remains for joint-actuated underwater robots. These systems rely on actuators that do not rotate continuously and include single-DOF mechanisms as well as multi-joint manipulators or locomotion modules. Commercial hardware currently offers either relatively low-cost underwater servos, suitable as building blocks for the development of custom prototypes but typically limited to position control without access to velocity or current feedback, or industrial-grade underwater manipulators and grippers that are prohibitively expensive and proprietary, providing very little flexibility for development and integration. This lack of accessible and open solutions significantly limits the development of lightweight, modular, and affordable underwater robots intended for scientific research. In practice, it often prevents academic groups from achieving field-ready prototypes and hindering broader community-driven innovation.

Open-source robotic toolkits have significantly impacted robotics research and development by providing accessible, modular, and customizable platforms. Examples include the Soft Robotics Toolkit for bio-inspired actuators \cite{holland2014soft}, the TurtleBot series for autonomous navigation \cite{amsters2020turtlebot}, and the Poppy Project for humanoid robotics \cite{lapeyre2014poppy}. These toolkits have played a crucial role in advancing robotics education and research by lowering entry barriers and fostering community-driven innovation. However, while many open-source robotic frameworks exist for terrestrial and aerial applications, open-source hardware solutions for underwater robotics remain scarce, particularly in the domain of actuation and manipulation.

In this work, we propose a cost-effective and modular toolkit for developing underwater robotics systems such as manipulators, grippers, or any other innovative design requiring waterproof actuation. The toolkit is the result of several years of experience in the development and deployment of underwater robotic platforms in real conditions. It aims to reduce development time and mitigate the risks associated with underwater robotic prototyping by offering open, off-the-shelf solutions for waterproofing, electronics, control, and software integration. Designed to be relatively low-power and low-cost, the toolkit offers an accessible path for research groups to rapidly develop and test underwater robotic systems suitable for field conditions with minimal custom hardware design.

The key contributions of this paper are:

\begin{itemize}
\item  The design of an Underwater Robotic Joint (URJ), tested up to 40 m depth, comprising a waterproof motor canister and an early-leakage detection system.
\item  A compact control and power management system for actuators control and data acquisition from a set of common sensors.
\item  A ROS\,2-based control framework integrating sensing and actuation.
\item  Field demonstration of the toolkit functionalities with application examples derived from the authors' extensive experience on marine field operations.
\end{itemize}

By making these designs publicly available under a \textit{Creative Commons license}, we aim to (1) lower the barrier to entry for researchers and developers working in underwater field robotics, (2) improve reproducibility of experimental results, fostering greater collaboration in the field, and (3) encourage community-driven improvements, leading to more robust and innovative designs.

The remainder of this article is structured as follows. Section II reviews relevant related work. Section III details the architecture and components of the proposed toolkit, while Section IV showcases three field-tested applications demonstrating its use in real underwater scenarios. Section V discusses the results and implications of the approach, and Section VI concludes the paper by outlining future directions.

\section{RELATED WORKS}
Existing underwater robotic platforms encompass a variety of vehicles and manipulation systems. Many rely on custom-built or industrial-grade actuators that require specialized waterproofing, often using hydraulic or high-performance electric drives. While hydraulic actuators dominate heavy-duty subsea operations due to their high power density and resistance to pressure variations \cite{moore2010underwater}, they are less suitable for precision operations and come with increased complexity, weight, and maintenance challenges. Conversely, electric actuators offer more compact, energy-efficient alternatives, making them suitable for lighter, more modular robots—a growing trend in research-oriented and field-deployable systems.

Among vehicle propulsion and manipulation technologies, propellers and robotic joints serve distinct functions. Propellers and thrusters, provide primary mobility for AUVs and ROVs. These have been well established in the open-source community, with robust integration into platforms such as ArduSub \cite{ArduSub2025}. On the other hand, robotic joints and manipulators require precise control and feedback, often demanding specialized underwater-rated actuators. Interest in these types of systems is increasing due to the growing importance of underwater manipulation and bioinspired solutions. Research into underwater manipulation has focused on various gripper designs \cite{mazzeo2022marine} and robotic arms \cite{sivvcev2018underwater}. Similarly, research on underwater legged robotics \cite{picardi2023underwater} and other bio-inspired platforms has proposed different limb design concepts \cite{aguzzi2021research}. While these systems offer manipulation capabilities and field use has been reported for several prototypes, open-source and open-hardware solutions remain scarce.

Companies such as \textit{Schilling Robotics}, \textit{Saab Seaeye}, \textit{Oceaneering}, \textit{Deep Trekker}, and \textit{Reach Robotics} manufacture high-performance, depth-rated manipulators. However, these systems remain proprietary and largely inaccessible to many researchers, with prices starting at 10{,}000\,EUR for a small manipulator with three functions (degrees of freedom) and a relatively shallow depth rating, and reaching several hundred thousand euros for more advanced models.

While the open-source movement has significantly contributed to advancing the field of underwater robotics, with initiatives focusing on vehicle platforms, either ROVs \cite{BlueROV2_2025} or AUVs \cite{LoCOAUV2021,MeCOAUV2025}, and control or simulation software \cite{stonefish, Manhaes_2016}, practically no efforts have been dedicated to open-hardware underwater manipulators and grippers. Existing open-source frameworks have democratized access to underwater robotics, enabling researchers and developers to build and operate underwater robots at lower costs. For example, in the last ten years, almost a hundred publications have utilized or modified BlueROV \cite{BlueROV2_2025} for research works on control, autonomy, sensing, and environmental monitoring. Taken together, these efforts highlight the progress in open-source underwater vehicles but also reveal a critical gap in accessible, open-hardware solutions for underwater joint driven robots, which the present toolkit is designed to address.

\section{TOOLKIT COMPONENTS}
This section describes the set of hardware and software modules of the proposed toolkit, covering the mechanical design of the underwater robotic joint, the control and power management electronics, and the corresponding ROS2 software stack.
\subsection{Underwater Robotic Joint}
\begin{figure}[htbp]
    \centering
    \includegraphics[width=0.45\textwidth]{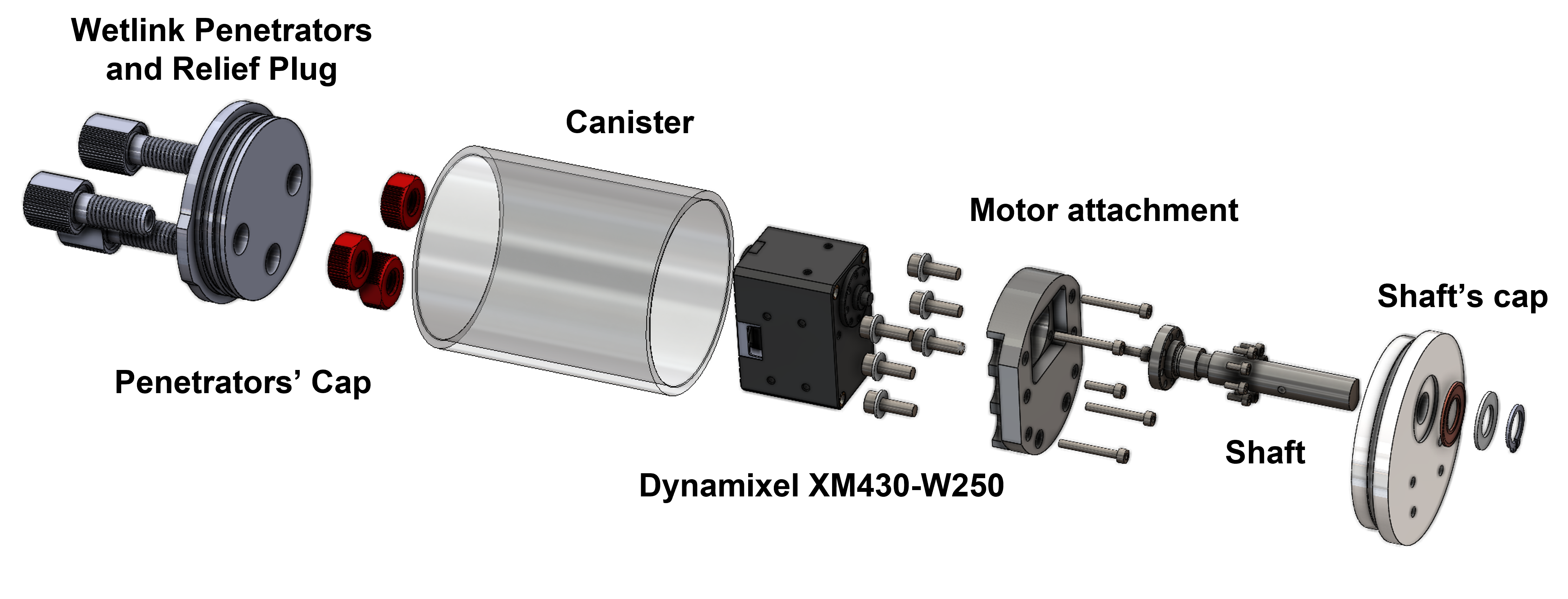} 
    \caption{Exploded view of the Underwater Robotic Joint assembly. Wiring and DHT11 humidity sensor are not shown}
    \label{fig:uwrj_assembly} 
\end{figure}

The Underwater Robotic Joint (URJ) consists of a sealed aluminum canister enclosing a Dynamixel XM430-W350 smart servo motor and a capacitive humidity sensor and a thermistor (DHT11) for early leakage detection as shown in Fig.~\ref{fig:uwrj_assembly}. The motor selection was dictated by the wide use of Dynamixel  motors, allowing users to choose from position, velocity, and current control modes, and providing great versatility for research projects involving both locomotion and manipulation tasks. This section will go through the mechanical design of the canister, lids and assembly components, the description of the early leakage detection system, and the hyperbaric chamber testing to validate the sealing and structural resistance under pressure.

\subsubsection*{Mechanical design and fabrication}
The URJ is implemented as an aluminum cylindrical housing (diameter 66 mm, height 82 mm, thickness 3 mm) that accommodates the motor and interfaces with a front and a rear cap, forming the basis for the shaft transmission, cable routing, and sealing strategy described below.

The motor drives a shaft penetrating the front cap, while the Wetlink penetrators in the rear cap carry power, serial communication with the motor, and the feedback from the DHT11 for early leakage detection. An additional Relief Plug is added to the rear cap to equalize the pressure during the assembly phase. This solution maintains atmospheric pressure inside the canister. 
The cylinder is manufactured by cutting individual canister sections from a longer extruded tube of 6061 T6 aluminum, chosen for corrosion resistance and machinability, with thickness (3 mm) dictated by buckling limits under combined axial and radial loads (assumptions of fixed ends and thin walls for the equivalent of 200 m depth, \cite{weingarten1968buckling}). The shaft is made of 316 stainless steel to prevent corrosion. The front cap is made of ABS to facilitate strong adhesion to aluminum. 
A retaining ring was added right outside of the front cap to prevent axial pressure from damaging the motor and improve the sealing and overall structural integrity. The front cap is permanently glued to the cylinder with a hybrid-polymer adhesive sealant (Multifiss, high-adhesion, moisture-curing) selected for its strong bonding to dissimilar materials and proven resistance to UV exposure, humidity, and marine environments. The rear cap is removable and it features two grooves dimensioned to host two O\!-rings (UNI O\!-ring standard OR2225: cross section 1.78 mm, internal diameter 56.87 mm). The cap lip thickness was selected to withstand external shear stress. Finally, the dynamic sealing on the motor shaft was implemented with a single groove with two O\!-rings (UNI O\!-ring standard OR106: cross section 1.78 mm, internal diameter 6.75 mm). The resulting dry weight of the URJ including shaft, motor, end caps and penetrators is 449 g. When submerged, the UJR has negative buoyancy corresponding to a underwater weight of 250 g. The unitary cost of manufacturing excluding externally purchased plugs, motor and humidity sensor, and 3D printed components was around 81€ for a batch of 21.

\subsubsection*{Early Leakage Detection System}
\begin{figure}[htbp!]
    \centering
    \includegraphics[width=0.45\textwidth]{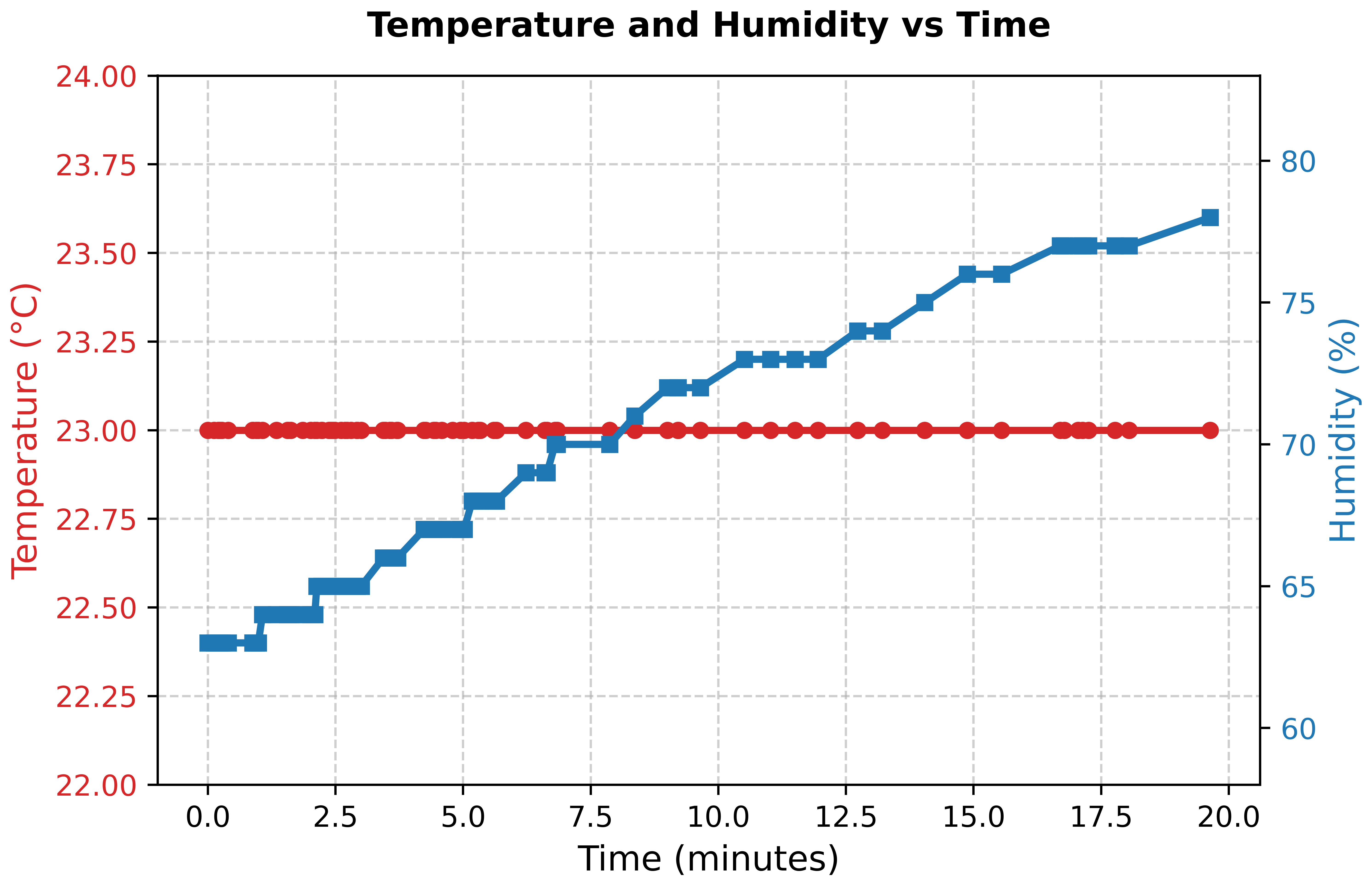} 
    \caption{Response of the humidity sensor following water ingress into the canister at T=0. The plot shows the evolution of temperature and humidity over time, highlighting the sensor’s reaction to the presence of moisture.}
    \label{fig:early_leakage_test} 
\end{figure}

The early detection of leakage is of paramount importance for any underwater system. The most commonly adopted solution consists of sponge-tipped probes that short-circuit two electrodes when the probe becomes wet. While this is a practical and low-cost solution, leakage is only detected once the sponge is soaked, which may be too late to prevent damage to the system. In practice, leakages often begin with just a few droplets of water. To detect leaks as early as possible, our solution uses a capacitive humidity sensor and a thermistor (DHT11), typically employed in agricultural applications, employing a proprietary single-wire communication protocol. These sensors are still extremely inexpensive yet capable of detecting small variations in internal conditions, with a temperature resolution of 1$^\circ$C, a relative humidity resolution 1\%, and a sampling rate of around 0.5 Hz.
The ability of the proposed system to detect a small leakage was tested on a single URJ by inserting a drop of water in the canister through the plug vent using a pipette at t = 0. As depicted in Figure \ref{fig:early_leakage_test}, while the temperature remains constant after the leakage, the relative humidity begins to rise after about one minute and steadily increases for roughly 20 minutes, reaching the upper range of the sensor (from approximately 63\% to 80\% RH). The canister was kept fully static during the test, which explains the relatively slow humidity increase; under conditions involving motion, agitation, or vibration, the same droplet would be expected to saturate the internal volume much more rapidly.

\subsubsection*{Hyperbaric chamber testing}
\begin{figure}[htbp!]
    \centering
    \includegraphics[width=0.485\textwidth]{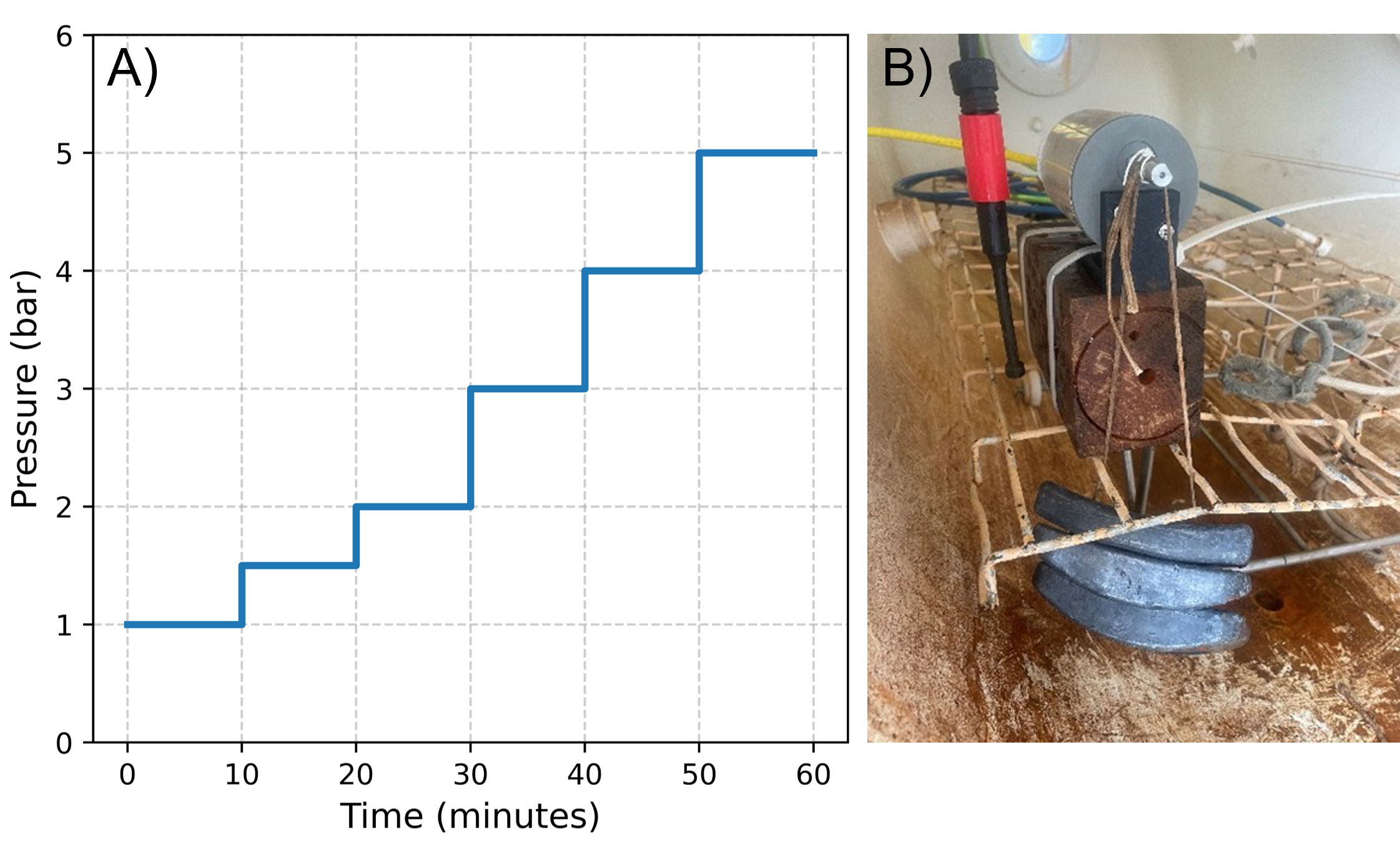} 
    \caption{A) Pressure levels inside the hyperbaric chamber. B) Underwater Robotic Joint under a static load.}
    \label{fig:hyperbaric_tests} 
\end{figure}

The static and dynamic sealing performance, as well as the structural integrity of the URJ under external pressure, were evaluated in a hyperbaric chamber (IBERCO model IB-80x150) following the protocol illustrated in Fig. \ref{fig:hyperbaric_tests}. The URJ was mounted on a custom frame fitting the chamber and subjected to progressively increasing static pressure levels in 1-bar increments (except for the first step, performed for safety) with each step lasting 10 minutes (Fig. \ref{fig:hyperbaric_tests}A). Absolute pressure was increased up to 5 bar, equivalent to the hydrostatic pressure at approximately 40 meters depth in seawater.

At each pressure step, the motor was left idle for the first 5 minutes and then commanded to rotate repeatedly from $-180^\circ$ to $180^\circ$ during the following 5 minutes. The same procedure was repeated while applying increasing transversal loads to the motor shaft to assess whether lateral forces could compromise dynamic sealing. Loads ranged from 0 to 8~kg of suspended lead weights, corresponding to approximately 0.91--7.28~kgf (8.9--71.4~N) of effective force underwater, applied by hanging individual 1~kg masses from the shaft as depicted in Fig.~\ref{fig:hyperbaric_tests}B.

Across all pressure levels and load conditions, no leakage was detected by the internal humidity sensor, and the motor maintained full functionality, indicating that both the sealing strategy and structural integrity of the URJ remained unaffected by the combined effects of external pressure and transversal loading.

\subsection{Control and Power Management Electronics}
The control and power management electronics constitute the functional equivalent of an autopilot system for propeller-driven vehicles—such as Pixhawk-class controllers or Raspberry-Pi-based autopilot boards used in the BlueROV2—providing the interfaces and I/O required for sensing, actuation, and onboard autonomy.
\begin{figure}[htbp]
    \centering
    \includegraphics[width=0.45\textwidth]{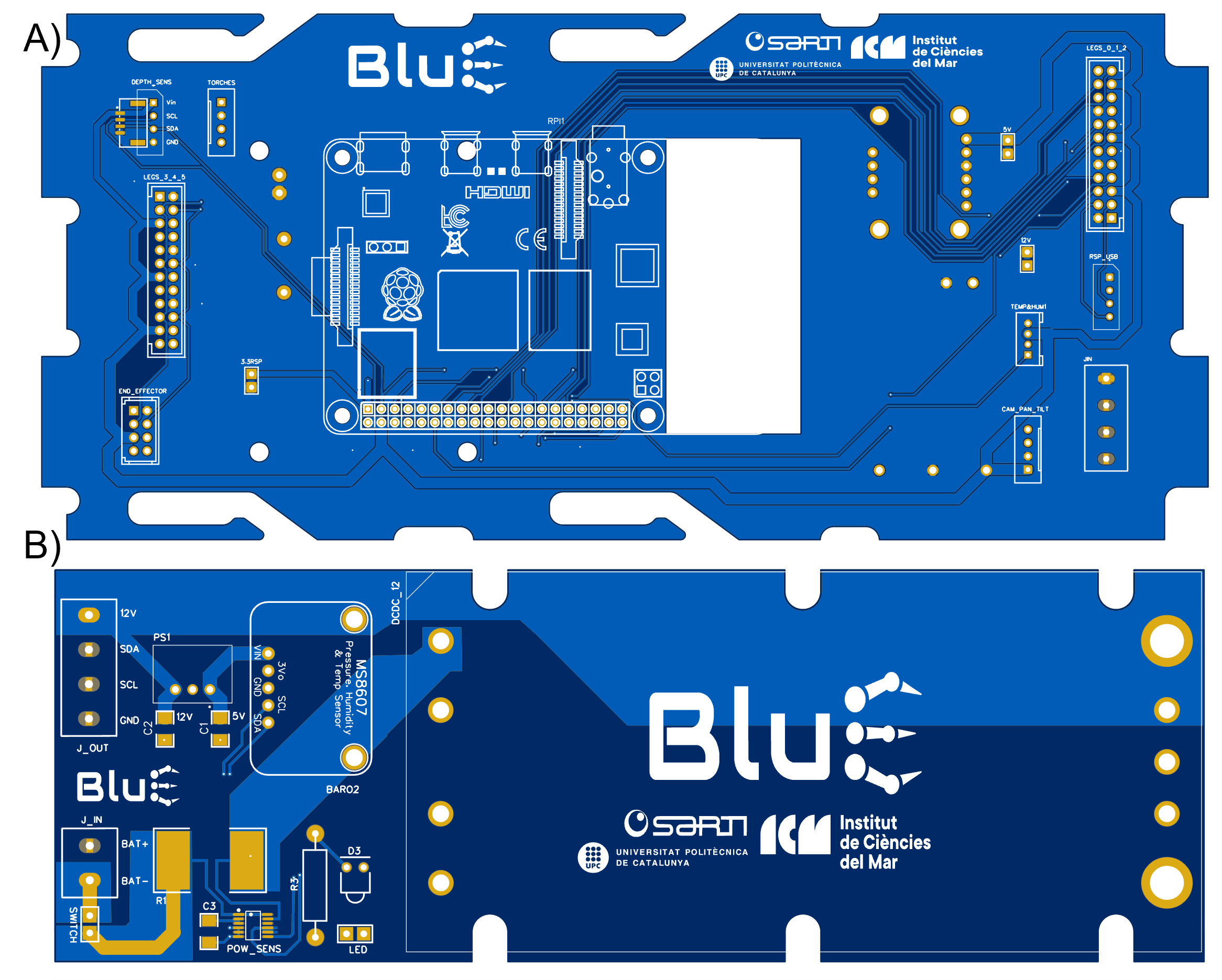} 
    \caption{A) PCB of the control electronics. B) PCB of the power management electronics.}
    \label{fig:pcb} 
\end{figure}

The design of the control electronics herein presented was guided by the following principles: 
\begin{itemize}
    \item Enable the core functionalities of an underwater vehicle/ manipulator including motor control, proprioception, monitoring of power consumption and leakages, tethered communication for topside monitoring and control; 
    \item Ensuring wide accessibility by using components with available free software and open-source libraries. 
\end{itemize}
The result is a system architecture based a Raspberry Pi4. The control board (Fig. \ref{fig:pcb}A) supports any actuator or motor driver that communicates via the RS-485 serial protocol, allowing users to adopt the actuators employed in this work or select among the many commercially available RS-485-based alternatives. Through the RS-485 link the user can send commands to the actuators, read their internal status, and receive feedback on their state including angular position, velocity, and current. 
In addition to motor control, the control board provides interfaces for a range of onboard sensors and actuators. General-purpose sensor integration is achieved through I\textsuperscript{2}C lines and dedicated GPIO pins, enabling the use of inertial, environmental, and power-monitoring devices, as well as additional modules as long as their addresses do not conflict with those already present on the bus. PWM channels directly controlled by the Raspberry Pi allow driving dimmable lights or auxiliary servomotors, while spare GPIO lines support further expansion for custom behaviors or instrumentation. The board accepts a regulated 12 V input and incorporates a high-current DC–DC converter to supply 5 V to the Raspberry Pi and peripheral sensors. A temperature and humidity sensor identical to that used in the URJ is integrated for early leak detection. Tethered communication for topside monitoring and control is implemented by integrating the \textit{Blue Robotics} Fanthom X Ethernet interface for long-distance communication. Specific devices used in this implementation include an IMU, a pressure and humidity sensor suite, a power-monitoring module, and a single-wire temperature–humidity probe (BNO055, MS8607, LTC2945, and DHT11, respectively).

The power management board (Fig. \ref{fig:pcb}B) complements this architecture by regulating the power distribution to motors, sensors, and communication modules. It accepts input voltages up to 24 V (in this work supplied by a 12 V, 25 Ah Li-Po battery) and delivers a stable 12 V rail to the control board. The motors represent the largest energy load and include an integrated overcurrent protection mechanism that prevents prolonged high current draw by temporarily shutting off the motor when overloaded. Consequently, the only additional overcurrent safeguard on the board is an 18 A fuse, while the integrated DC–DC converter (VICOR V28A12C200BL) can supply up to 18 A on the output line. The board also hosts an internal I\textsuperscript{2}C bus for monitoring environmental conditions inside the watertight enclosure, including pressure, temperature, and humidity (MS5837).

Both boards are designed to fit in standard watertight enclosures of inner diameter 100 mm and length 300 mm.

\subsection{ROS2 Software Stack}
A ROS2 software stack has been developed to interface with all components of the toolkit, providing modular control of underwater robotic joints, sensors, and communication devices. The system leverages ROS2 for distributed execution, scalability, and compatibility with standard robotic frameworks. The software architecture is organized into dedicated nodes responsible for sensor data acquisition and publication, and a motor control and state feedback subsystem built upon the \textit{ros2\_control} framework.

Sensor nodes follow a uniform design strategy in which each module handles a single sensor, abstracts the underlying communication protocol, and publishes standardized ROS2 messages whenever possible. Publicly available sensor drivers were directly reused or adapted into ROS2 nodes, and the corresponding implementations are documented in the associated GitHub repository \cite{Toolkit2025}. This modular approach allows users to integrate additional devices provided their communication interfaces (typically I\textsuperscript{2}C or GPIO) are supported by the control electronics. Custom messages are only introduced when required, such as for the DHT11 temperature and humidity sensor, which relies on a proprietary single-wire protocol. The sensors integrated in this work include internal environmental monitoring, inertial measurement, power monitoring, and external pressure and depth estimation.

Actuation is handled through a dedicated hardware interface based on \textit{ros2\_control}. The implementation is built upon the public Dynamixel hardware plugin for RS-485 communication, which has been extended in this work to include a simple transmission mechanism supporting gear reduction, actuation direction, and joint offsets. The framework enables position, velocity, and current control modes, and provides joint state feedback (position, velocity, effort) through standard ROS2 interfaces. If alternative actuators are selected, the corresponding hardware interface must be integrated following the \textit{ros2\_control} guidelines similarly to the modular structure employed here.

Overall, the software stack enables robust sensing, communication, and multi-mode actuation for underwater robotic platforms, while maintaining extensibility through ROS2 conventions, modular node design, and reuse of open-source drivers.

\section{APPLICATIONS OF THE TOOLKIT}
\begin{figure}[htbp]
    \centering
    \includegraphics[width=0.45\textwidth]{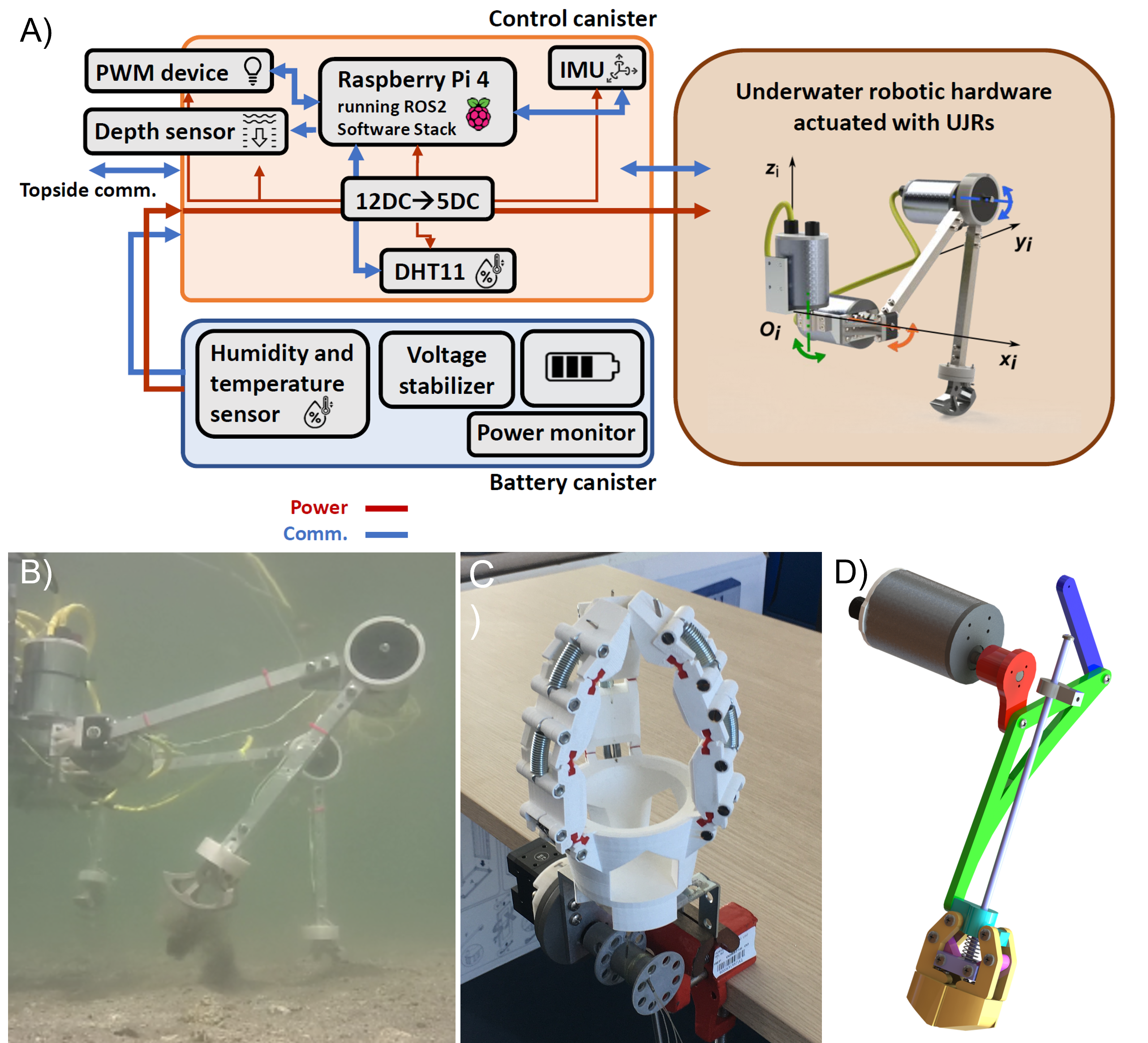} 
    \caption{A) System-level block diagram of the toolkit, illustrating the interconnection between the control and power management electronics, ROS 2 software stack, and the Underwater Robotic Joints (URJs). Power and communication links in the battery canister are not shown. B) RRR serial robot. C) Tendon driven soft manipulator. D) Underactuated sampling system}
    \label{fig:toolkit} 
\end{figure}

The building blocks of the proposed toolkit have been used to design, develop, and deploy multiple underwater robotic prototypes following the schematics reported in Fig. \ref{fig:toolkit}A. This section presents three representative applications: (i) a three-degrees-of-freedom (3-DoF) RRR serial manipulator employed as a leg in a field-tested underwater legged robot, (ii) a tendon-driven soft robotic gripper, and (iii) an underactuated sediment sampling mechanism. All systems were actuated exclusively using the URJ and control electronics described earlier, demonstrating the modularity and reusability of the toolkit across diverse morphologies and tasks.

\subsection{RRR Serial Manipulator}
Multiple URJs can be combined to form articulated kinematic chains. The first implementation of the toolkit was a 3-DoF serial manipulator (Fig. \ref{fig:toolkit}B) consisting of three rotational joints (RRR) used as a leg of the underwater legged robot SILVER2 \cite{picardi2020bioinspired}. The three joints—coxa, femur, and tibia—were assembled by combining URJs with custom metallic and polymeric link interfaces. The tibia shaft was connected to a torsional serial elastic actuator to support compliant interaction with the seabed, and a modular foot was mounted at the end effector.

\begin{figure}[htbp]
    \centering
    \includegraphics[width=0.4\textwidth]{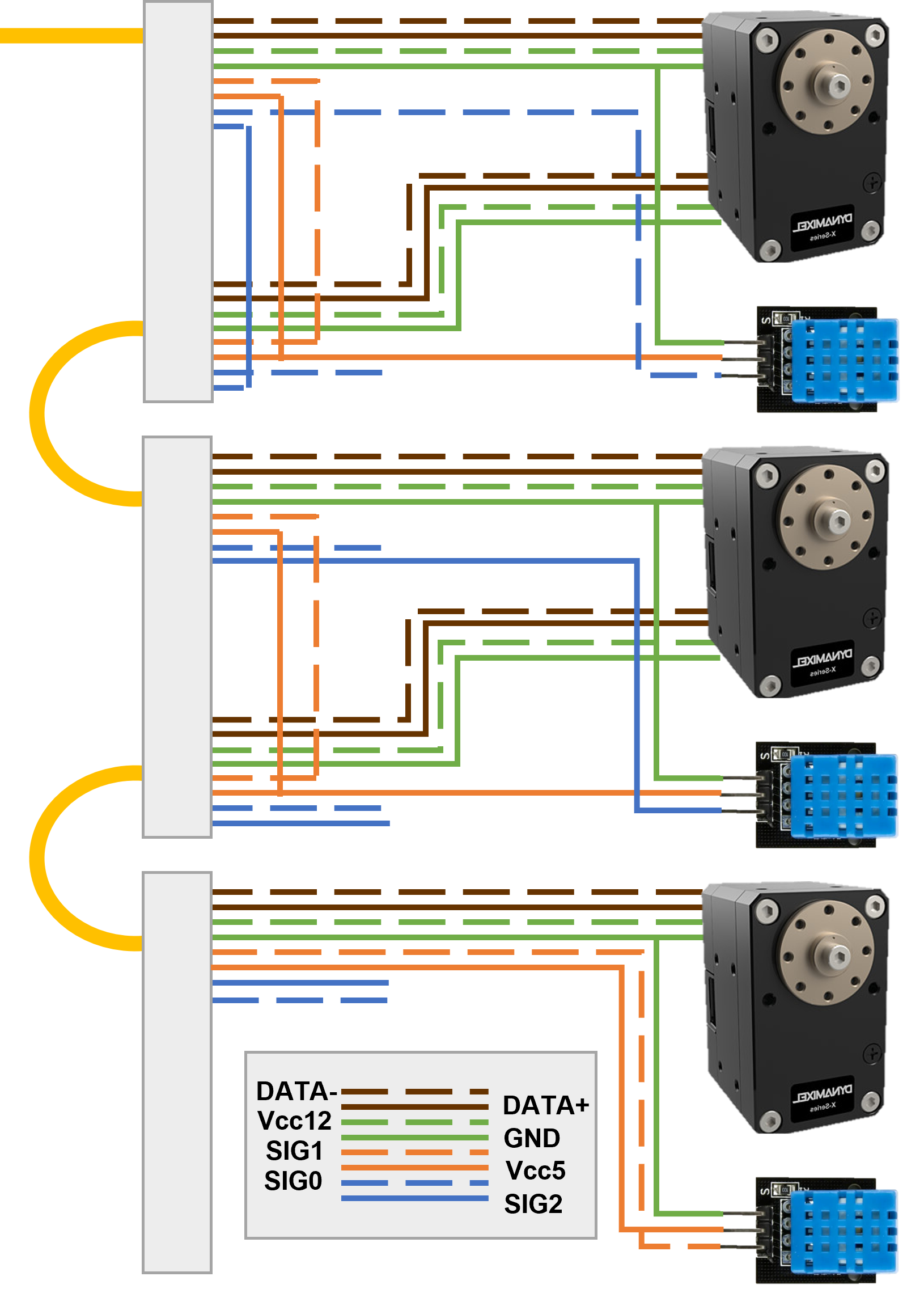} 
    \caption{Wiring scheme for actuation and early detection system in the RRR serial manipulator developed with presented toolkit.}
    \label{fig:leg_wiring} 
\end{figure}

To enable daisy-chaining of actuators and acquisition of humidity measurements, the first two joints employ three penetrators each—one vent plug and two penetrators fitted with an 8-conductor underwater cable—whereas the distal joint uses two penetrators (one vent plug and one cable). The wiring scheme for actuation and early leakage detection is illustrated in Fig.~\ref{fig:leg_wiring}. The 8-conductor cable allocates four lines to motor power and communication (\textit{DATA-, DATA+, V\textsubscript{cc}12, GND}), one line to supply the DHT11 sensor at 5~V (\textit{V\textsubscript{cc}5}), and the remaining three lines to stack up to three URJs and independently acquire the single-wire signals from their respective DHT11 sensors (\textit{SIG0, SIG1, SIG2}).

\subsubsection*{Field Deployment}
\begin{figure}[htbp!]
    \centering
    \includegraphics[width=0.49\textwidth]{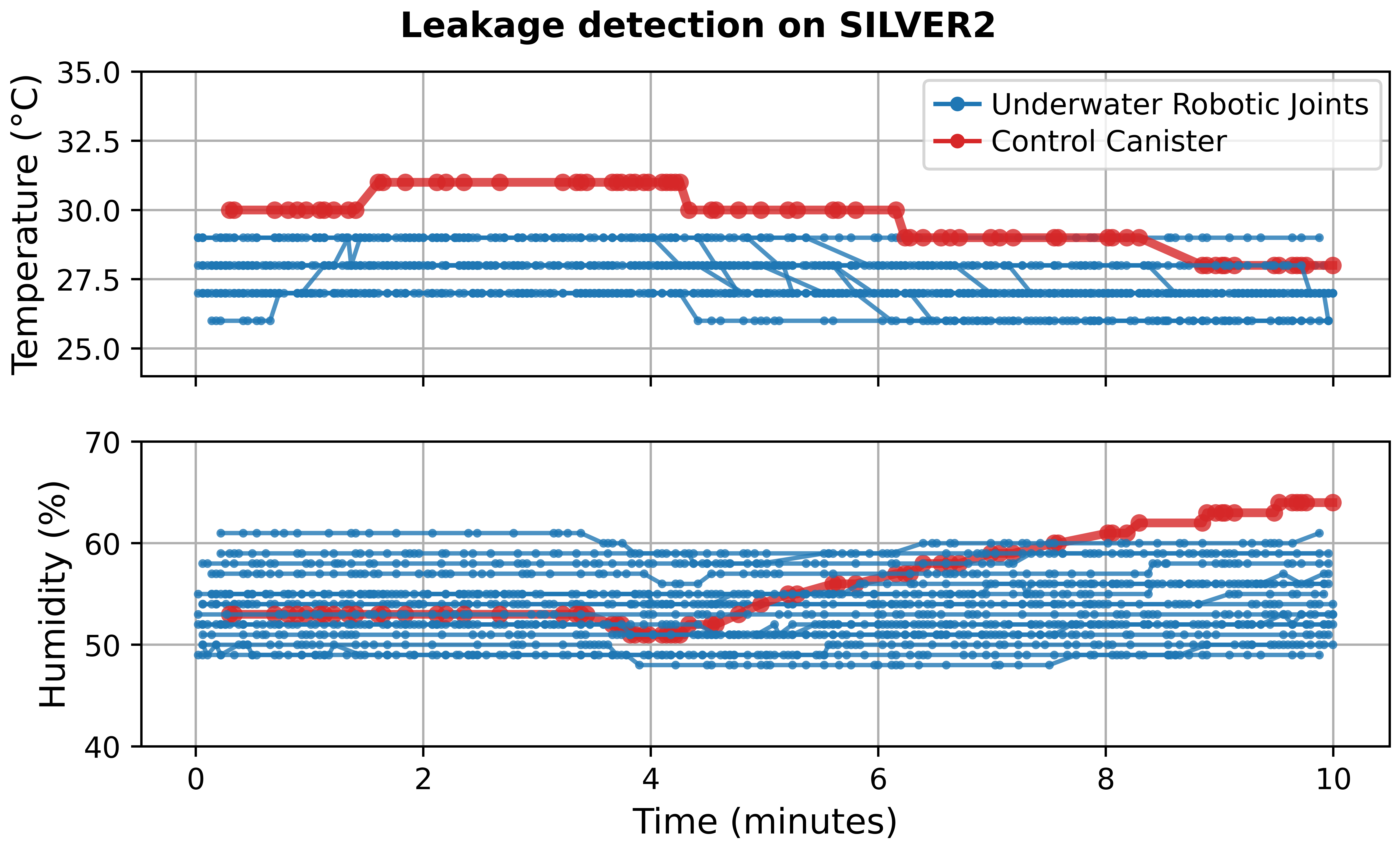} 
    \caption{Response of the humidity sensors array in the SILVER2 robot following water ingress into the control canister at T= 3 min 27 s. The plot shows the evolution of temperature and humidity over time for the motor canisters (blue) and control canister (red).}
    \label{fig:leakage_test_silver} 
\end{figure}

\begin{figure}[htbp!]
    \centering
    \includegraphics[width=0.45\textwidth]{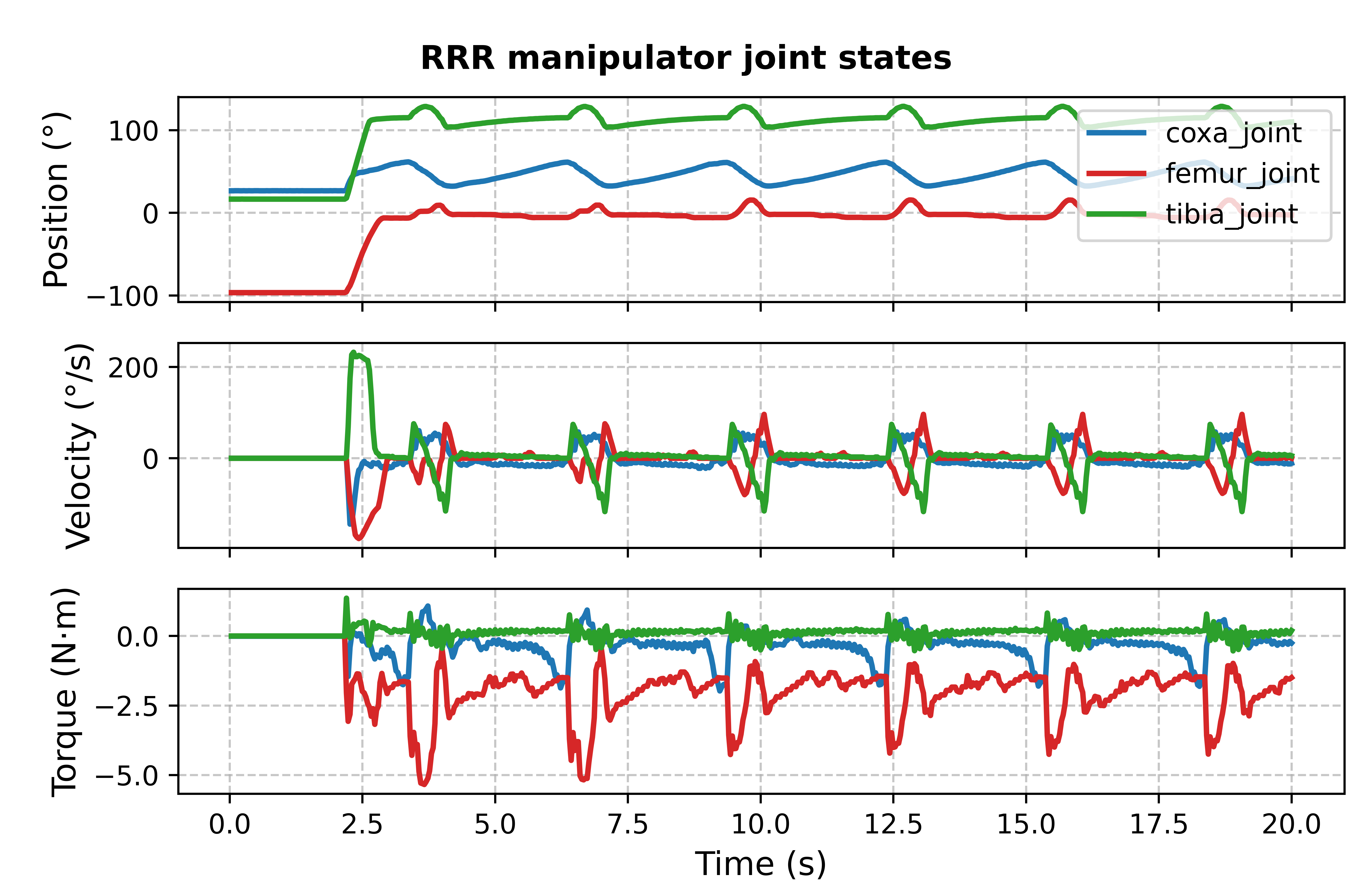} 
    \caption{Joint state feedback from the three Underwater Robotic Joints of the RRR-manipulator during position-control (inverse kinematics) while tracking a semi-elliptical end-effector (foot) trajectory as part of a gait cycle.}
    \label{fig:joint_states_silver} 
\end{figure}

The leg has undergone extensive laboratory and field testing, operating continuously for several hours over multiple experimental sessions, thoroughly reported in the corresponding scientific literature, without failure. The system achieved a validated operational depth of 30 m, with stable actuator performance and no water ingress. Prior to the introduction of the early leakage detection system, water ingress during field testing was detected at approximately 45 m depth in one canister right before the system shutdown via the commonly used sponge-tipped probes, triggering the immediate recovery of the platform. Water then propagated to the adjacent canisters through the internal wiring. This scenario underscores the importance of the integrated early-leakage system and demonstrates realistic fault behavior in field conditions. Figure \ref{fig:leakage_test_silver} reports the response from the early leakage detection system for a simulated water ingress in the control canister of the Underwater Legged Robot. During a submerged test at 1 m depth with one screw intentionally removed from the control canister cap, the onboard humidity sensor detected a humidity rise less than one minute after ingress began at T = 3 min 27 s. The robot was retrieved 5 minutes later, and roughly 20 mL of water was found inside the canister. Note that the measured baseline relative humidity (RH) inside many canisters was relatively high (often above 50--60\%), since the tests were performed during summer months when ambient humidity was elevated. For this reason the early-warning criterion implemented in our tests is based on the change in RH relative to the initial condition rather than on an absolute RH threshold.
Joint state feedback during position-control inverse kinematics is reported for one leg in Fig. \ref{fig:joint_states_silver}; the controller drives the joints to follow a semi-elliptical trajectory of the end effector (foot) as part of a cyclic gait.

This application validates the use of URJs for precise, multi-joint actuation in realistic underwater environments, including locomotion tasks requiring repeated range-of-motion cycles under load.

\subsection{Tendon-Driven Soft Gripper}
Cable- or tendon-driven systems offer several advantages for underwater robotics, including lightweight distal structures, efficient remote force transmission, and natural compatibility with soft materials to create adaptable, underactuated manipulators \cite{jeong2021reliability}. Leveraging these properties, we developed a tendon-driven soft gripper actuated by a single URJ (Fig. \ref{fig:toolkit}C).

A custom spool is rigidly mounted to the URJ shaft and drives three tendons in parallel, each corresponding to one finger. The gripper consists of a hollow palm, connected to the URJ via an aluminum L-connector, and three identical bioinspired fingers. Each finger has three rigid phalanges connected by soft silicone joints, allowing passive adaptation to irregularly shaped objects and providing an inherent limit on maximum gripping force.

\subsubsection*{Field Deployment}
The gripper was evaluated through controlled tank tests and multiple field deployments, operating reliably up to a maximum depth of 30 m without exhibiting any water leakage. Across these trials, the system successfully executed complete pick-and-place sequences for a wide range of objects, including a plastic bottle, plastic bag, a 5 cm-radius ball, a small tin can, and a fragment of fishing net \cite{picardi2024seabed}. These objects span rigid, deformable, and flexible categories, demonstrating the adaptability of the tendon-driven soft design. Failures occurred only with highly deformable or extremely lightweight items (e.g., paper tissue, label) or with objects whose dimensions exceeded the effective grasping envelope (e.g., large plastic bag, large plastic jug). Importantly, these limitations are inherent to the soft finger geometry and compliance characteristics, not to the underlying actuator or waterproofing provided by the toolkit.

Across deployments, the gripper demonstrated reliable underwater operation and consistent force transmission. When a stable grasp is achieved, the system can exert a maximum pull-out force of 37 ± 3.7 N, sufficient to lift nearly 4 kg under water \cite{picardi2023user}, highlighting its capability for both delicate and load-bearing manipulation tasks.

This application showcases the ability of the toolkit to support tendon-driven, soft, and bioinspired underwater systems without any modification to the base actuation unit. The same URJ module used for serial manipulation and underactuated mechanisms can directly actuate compliant grippers, emphasizing the flexibility and extensibility of the proposed open hardware platform.

\subsection{Underactuated Sediment Sampling Mechanism}
The development of a simple and compact robotic module capable of repeating sediment sampling and depositing collected material into an integrated storage compartment is of significant interest for advancing robot-based benthic sampling techniques. Such systems reduce reliance on large, specialized vehicles and enable lightweight platforms—including legged robots and small AUVs—to perform ecological and geochemical studies autonomously.

Building on this motivation, we designed an underactuated sediment sampling mechanism actuated entirely by a single URJ (Fig. \ref{fig:toolkit}D). The system is composed of:

\begin{enumerate}
    \item a four-bar linkage arm that transports the sampling unit between the seabed and the onboard storage box,
    \item a passive grab mechanism that closes during contact with the sediment and opens automatically during deposition, and
    \item an internal storage space that accumulates multiple samples during a mission.
\end{enumerate}

The URJ directly drives the crank-link of the four-bar mechanism, enabling a fully mechanical sequence of sampling, lifting, and deposition without additional actuators or sensors. This approach minimizes system complexity while ensuring reliable, repeatable motion underwater.

\subsubsection*{Field Deployment}
The sampler was field-tested in shallow water (around 2 m depth) as part of a microplastics analysis study carried out in a small touristic harbour in Livorno, Italy in 2022, where it performed multiple collection–deposition cycles without intervention \cite{astolfi2024marine}. Although deployed at low depth, the module relies on the same URJ design validated up to 30 m, confirming the robustness and transferability of the actuation and sealing system. The mechanism effectively collected loose sediment and released it into the onboard container, demonstrating the feasibility of minimal-actuation, mechanically robust sampling systems built using the proposed toolkit.

This application highlights how the toolkit enables the rapid development of underactuated, task-specific benthic tools, supporting ecological monitoring, environmental sampling, and field robotics research without increasing system cost or complexity.

\section{DISCUSSION}
\label{sec:discussion}

Unlike many open-source robotic toolkits primarily designed for education or introductory robotics, the proposed open toolkit addresses the needs of engineers and researchers developing \emph{field-deployable} underwater manipulators, bioinspired devices, and task-specific tooling. This section summarizes the advantages of the system, identifies current limitations and avenues for improvement, and highlights how the open-access nature of the toolkit supports reproducibility and community-driven development.

\subsection{Advantages over existing solutions}
The toolkit provides several advantages compared to both commercial underwater manipulators and existing open-source robotic platforms:

\begin{itemize}
    \item \textbf{Cost-effectiveness and accessibility.} Industrial underwater manipulators typically start above 10,000\,EUR for basic 3-DOF units and can reach several hundred thousand euros for advanced multi-function systems. Their closed-source nature additionally restricts modification and integration. Off-the-shelf underwater servomotors exist, but they typically fail to provide position, velocity and current feedback. In contrast, the proposed toolkit relies on commercially available components (e.g., Dynamixel servos, Blue Robotics penetrators) and standard machining or 3D-printing processes, significantly reducing cost, providing joint state feedback, and enabling fabrication in modestly equipped laboratories.

    \item \textbf{Early leakage detection system.} The URJ includes an early leakage detection system based on a capacitive humidity sensor and a thermistor (DHT11), originally used in precision agriculture for low-cost environmental monitoring. Compared to the commonly used sponge-tipped probes, which only trigger once the sponge is fully soaked and the electrodes short-circuit, our approach can detect leakage from a \emph{single droplet} of water through the rise in internal humidity. Although a systematic comparison between the two systems has not been carried out, the proposed solution produces significantly earlier warning and is particularly valuable in custom prototypes or research platforms where waterproofing is a primary source of risk and where early intervention can prevent actuator failure or electronic damage.

    \item \textbf{Modularity and reusability.} The same Underwater Robotic Joint (URJ) and control electronics were used to construct a 3-DOF RRR manipulator, a tendon-driven soft gripper, and an underactuated sediment sampler. This demonstrates the flexibility of the approach and the ability to rapidly reconfigure the system across manipulation and locomotion applications without redesigning the actuation module.

    \item \textbf{Novelty in the open-source ecosystem.} While the open-source movement has democratized access to underwater vehicles (e.g., BlueROV2 \cite{BlueROV2_2025}, LoCO AUV \cite{LoCOAUV2021}) and surface or terrestrial robots (e.g., TurtleBot \cite{amsters2020turtlebot}, the Soft Robotics Toolkit \cite{holland2014soft}), open-hardware solutions for underwater \emph{manipulation} remain scarce. To the best of our knowledge, the presented toolkit is among the first open-source, field-tested hardware frameworks specifically targeting underwater joints, manipulators, and end-effectors.

    \item \textbf{Field validation and robustness.} The toolkit has been successfully deployed in real underwater scenarios. The URJ and control electronics operated reliably during multi-hour missions, with the joint assemblies validated up to 30\,m depth without water ingress. The tendon-driven gripper demonstrated robust grasping performance on a variety of objects, and the underactuated sampling mechanism performed repeated collection cycles in shallow-water field trials. Further details on these deployments are provided in \cite{astolfi2024marine} and \cite{picardi2023user}.

    \item \textbf{ROS\,2 integration and extensibility.} The ROS\,2 software stack provides standardized interfaces, multi-mode actuation through \textit{ros2\_control}, modular sensor nodes, and compatibility with existing planning and control frameworks. Additional actuators and sensors can be integrated by extending or reusing the provided hardware interfaces.
\end{itemize}

\subsection{Limitations and recommended improvements}
Despite these advantages, the current toolkit exhibits several limitations that must be addressed in future revisions. 

\subsubsection*{\textbf{Depth rating}} The present canister design has been validated up to 40\,m in controlled hyperbaric chamber tests and up to approximately 30\,m during field deployments. Beyond this depth, leakage risk increases: during one field test, water ingress occurred at $\sim$45\,m in a single canister and subsequently propagated to neighbouring units through the cables. This failure was not caused by structural deformation of the aluminum canister—which provides sufficient thickness to avoid buckling at these pressures—but is most likely attributable to O\!-ring misplacement or imperfect seating during assembly. Given the sensitivity of the current sealing strategy to assembly precision, future improvements should include more robust O\!-ring retention features, redesigned penetrators, and pressure-compensated variants (e.g., oil-filled designs) compatible with immersion-rated motors.

\subsubsection*{\textbf{Oil-filling and motor compatibility}} Coreless brushed motors are not compatible with oil immersion, limiting depth-compensation options. Future designs should explore Brushless DC motors or oil-tolerant drivers to enable pressure-compensated operation.

\subsubsection*{\textbf{Lateral-load sensitivity}} In the existing configuration, the shaft relies on a single O\!-ring groove for dynamic sealing, which increases vulnerability to external lateral forces. As discussed earlier, external cantilever loads can deform the shaft path and compromise the seal. While testing in the hyperbaric chamber \ref{fig:hyperbaric_tests} under transversal static loads showed no leakage, impacts are more likely to cause water ingress. This limitation may be mitigated by introducing two separate O\!-ring grooves (at the cost of increased cap thickness), integrating external support bearings to reduce bending moments, or redesigning the front-cap assembly to better tolerate radial loading.

\subsubsection*{\textbf{Assembly complexity}} The current assembly requires numerous screws, multiple O\!-rings, adhesive bonding of the front cap, and careful alignment of seals and penetrators. While robust, this increases assembly time and elevates the risk of human error, especially in systems using multiple URJs (such as the RRR leg). Future revisions should consider reduced fastener counts, captive screws, dowel-based alignment features, and clearer assembly aids while preserving sealing performance.

\subsubsection*{\textbf{Torque and workload limitations}} Dynamixel XM-series actuators are well suited for research and general-purpose motion tasks but are not optimized for high-torque or long-duration load-bearing applications. For heavier manipulation tasks, the canister geometry may be adapted to accommodate brushless DC motors with higher efficiency.

\subsubsection*{\textbf{Biofouling and sediment exposure}} The toolkit underwent significant field testing, however continuous submerged time never exceeded 8 hours. Extended deployments in real marine environments will expose the system to biofouling, corrosion, and sediment abrasion. Future work should quantify these effects and evaluate surface treatments (e.g., anodizing, marine coatings) and anti-fouling strategies.

\subsubsection*{\textbf{Propagation of faults through wiring}} During extreme hyperbaric testing, water ingress in one canister propagated through shared wiring to adjacent modules. Isolated wiring harnesses, localized leak detection zones, and improved cabling topology will help mitigate this risk.

\subsection{Open access, reproducibility, and community development}
A core goal of this work is to make underwater manipulation research more accessible. All hardware and software designs—including CAD assemblies, CNC-ready models, PCB sources, BOMs, wiring diagrams, and ROS\,2 packages—are openly available at the following anonymized Github repository \cite{Toolkit2025}. This allows users to fabricate the system locally using standard machining facilities, PCB manufacturers, and low-cost 3D printers, or to modify the source files to accommodate custom actuators, sensors, or mechanical interfaces.

The open-hardware release supports:
\begin{itemize}
    \item reproducibility of experimental results;
    \item transparent evaluation and comparison across laboratories;
    \item community-driven improvement of the design;
    \item reduced development time for underwater field robotics.
\end{itemize}

\section{CONCLUSION AND FUTURE WORKS}
We presented a modular and open-access toolkit for underwater robotics that integrates waterproof actuation, power management, early leakage detection, and a ROS\,2-based control framework. The system has been validated in both laboratory and field conditions, through the implementation of multiple prototypes (RRR serial manipulator, tendon-driven gripper and underactuated sampler) demonstrating reliable operation up to 30\,m depth and confirming the versatility and robustness of the proposed components. By openly sharing all mechanical designs, electronic schematics, software packages, and fabrication files, we provide a reproducible and extensible platform intended to lower the barrier to entry for underwater field robotics research and development.

Future engineering work will focus on improving the sealing strategy and assembly robustness, including enhanced dynamic shaft sealing, redesign of O\!-ring retention features, and the development of pressure-compensated variants to extend the operational depth range. Additional efforts will explore higher-torque actuation options, such as brushless DC motors or alternative transmission mechanisms, to increase payload capacity and broaden the range of possible applications.

From a systems perspective, future work will investigate improved proprioceptive sensing (e.g., joint torque or inline force measurements), long-duration field deployments to assess biofouling and environmental durability, and further optimization of the ROS\,2 control stack, including characterization of real-time communication performance under realistic mission conditions. 

By releasing the toolkit under an open license and providing CNC-, PCB-, and print-ready files, we invite the underwater robotics community to adopt, modify, and extend the platform. Collaborative development will accelerate the transition from laboratory prototypes to robust, field-deployable systems for marine science, environmental monitoring, and subsea industrial applications.

\section*{ACKNOWLEDGMENT}
The present research was carried out within the framework of the 
activities of the Spanish Government through the “Severo Ochoa Centre of Excellence” granted to ICM-CSIC (CEX2024-001494-S) and the Research Unit Tecnoterra (ICM-CSIC/UPC). Projects that supported the work were the EU DIGI4ECO [grant number GAP-101112883], REDRESS [grant number 101135492], SUN-BIO [grant number 101157493-GAP-101157493], and MERLIN [grant number GAP-01189796].  Other participating actions are those funded by the Spanish Government (Ministerio de Ciencias, Innovación y Universidades-MICIU): AI4SEA (AIA2025-163346-C44) and SMART-ME (PID2024-1553440B-C31). G.P was partially funded by the European Commission through the HORIZON-MSCA-2021-PF-01-01, grant no. 101061354.

\bibliographystyle{elsarticle-num} 
\bibliography{biblio}



\vfill\pagebreak

\end{document}